%% file: main.tex
\pdfoutput=1

\documentclass[11pt]{article}

\usepackage[preprint]{acl/acl}

\usepackage[shortlabels]{enumitem}

\usepackage{times}
\usepackage{latexsym}

\usepackage[T1]{fontenc}

\usepackage[utf8]{inputenc}

\usepackage{microtype}

\usepackage{inconsolata}

\usepackage{graphicx}

\input{packages}

\title{LLM Web Dynamics: Tracing Model Collapse in a Network of LLMs}

\author{Tianyu Wang$^{\dagger}$ \\
Johns Hopkins University \\
\And
Akira Horiguchi \\
University of California, Davis
\AND
Lingyou Pang \\
University of California, Davis
\And
Carey E. Priebe \\
Johns Hopkins University
}

\newcommand\blfootnote[1]{%
  \begingroup
  \renewcommand\thefootnote{}\footnote{#1}%
  \addtocounter{footnote}{-1}%
  \endgroup
}

\begin{document}
\maketitle

\blfootnote{\\
$ ^{\dagger} $ corresponding author: \texttt{twang147@jhu.edu}.
}

\input{text/abstract}
\input{text/intro}
\input{text/related}
\input{text/methods}
\input{text/results}
\input{text/discussion}
\input{text/limitations}

\subsection*{Acknowledgments}
This work was supported by Air Force Office of Scientific Research (AFOSR) Complex Networks award number FA9550-25-1-0128, and Defense Advanced Research Projects Agency (DARPA) Artifical Intelligence Quantified award number HR00112520026.\\
We would like to thank Avanti Athreya, Youngser Park, Edward L. Wang, and Hayden Helm for their helpful feedback and discussions throughout the development of this manuscript.

\bibliography{citation}

\appendix
\input{text/appendix}
\newpage
\input{text/appendix2}

\end{document}

%% file: packages.tex
\usepackage{caption}
\usepackage{amsthm}

\usepackage{subcaption}
\usepackage{graphicx}

\usepackage{natbib}

\usepackage{amsfonts,amsmath,amssymb}
\usepackage{bbm}
\usepackage{xspace}
\usepackage{mathtools}

\usepackage{multicol}
\usepackage{enumitem}

\usepackage{algorithm}
\usepackage{algpseudocode}

\usepackage{wrapfig}

%% file: text/abstract.tex
\begin{abstract}
The increasing use of synthetic data from the public Internet has enhanced data usage efficiency in large language model (LLM) training. However, the potential threat of model collapse remains insufficiently explored. Existing studies primarily examine model collapse in a single model setting or rely solely on statistical surrogates. In this work, we introduce LLM Web Dynamics (LWD), an efficient framework for investigating model collapse at the network level. By simulating the Internet with a retrieval-augmented generation (RAG) database, we analyze the convergence pattern of model outputs. Furthermore, we provide theoretical guarantees for this convergence by drawing an analogy to interacting Gaussian Mixture Models.
\end{abstract}

%% file: text/intro.tex
\section{Introduction}

Synthetic data is crucial for training and scaling large language models. On the one hand, neural-scaling law assumes model performance will improve as the amount of training data increases \citep{kaplan2020scalinglawsneurallanguage}; on the other hand, the pool of human-generated data is approaching its natural limit \citep{villalobos2024rundatalimitsllm}.  Augmenting the training data with synthetic text has therefore become a common remedy \citep{jimaging8110310}. However, recent evidence suggested that the usage of synthetic data can lead to issues including parameter degradation \citep{dohmatob2024strongmodelcollapse}, amplified  biases \citep{wyllie2024fairnessfeedbackloopstraining}, and, in the most extreme cases, model collapse: a process characterized by the loss of information about the raw data distribution when an LLM is iteratively trained primarily on its own previous outputs \citep{shumailov2024}. Analyzing the process of model collapse, and more generally the perturbations from synthetic data, is thus essential for LLM training.

In our work, we introduce the \emph{LLM Web Dynamics} (LWD)—a framework that renders the full life-cycle of synthetic data \textit{tractable}, \textit{measurable}, and \textit{theoretically valid}.  
Our contributions are threefold:

\textbf{1.~Network-level model-collapse metric.}  
We design a network of interacting LLMs in which pretrained models from API communicate, receive opinions from each other, and update memory via retrieval-augmented generation (RAG). Individual responses are mapped into an embedding vector space and then converted to a pairwise distance matrix which quantitatively captures response similarities. As collapse unfolds and response similarity rises, this matrix shows drifting in geometry features accordingly, which provides a model-agnostic stability measuring metric. This metric reflects the stabilization of collective communicating dynamics of all participant LLMs.

\textbf{2.~Internet ecosystem simulator (API-only, cost-efficient).}  
We anticipate a near-future feedback loop where synthetic texts generated by LLM agents populate the Internet and are later reused as training data or prompt augmentations in downstream applications \citep{martínez2023combininggenerativeartificialintelligence,seddik2024badtrainingsyntheticdata}. Our LWD framework constructs a virtual reality that mirrors real-world, AI-rich Internet environment by deploying a centralized RAG memory updating algorithm. Since each agent operates solely through API-accessible models, the framework approximates incremental fine-tuning without the need for retraining or additional fine-tuning, while also enabling the exploration of more complex scenarios.

\textbf{3.~Theoretical grounding via a Gaussian Mixture Model (GMM) proxy.}  
To validate the collapse process of our LLM system analytically, we construct an analogue network of GMMs, motivated by \citet{wang2025}. For a fixed prompt, an LLM’s answers form a probability distribution, and GMMs serve as density approximators \citep{goodfellow2016}. This proxy offers theoretical guarantees for our claims and serves as a cost-efficient tool for gaining preliminary insights into LLM behavior at larger scales.

%% file: text/related.tex
\section{Related Work}

This phenomenon of model degradation is well documented and has been observed across domains, tasks, and model types. \citet{alemohammad2023selfconsuminggenerativemodelsmad} introduce self-consuming training loops and show that, without a constant infusion of fresh real data, models converge to “Model Autophagy Disorder” (MAD). In image generation, \citet{bohacek2025nepotisticallytrainedgenerativeaimodels} demonstrate that even a small share of self-generated images used during retraining triggers self-poisoning, reducing both realism and diversity. \citet{shumailov2024} first formalize this phenomenon as model collapse and demonstrates its existence in practice via GMMs, variational autoencoders (VAEs), and LLMs. 

Quantifiable and interpretable  model collapse have also attracted many attention in recent literature. Statistical surrogates are commonly used to provide instrumentally similar insights into this problem. In particular, regression-based methods are applied to simulate the iterative training of LLMs because they yield closed-form analytical risk expressions \citep{dohmatob2024strongmodelcollapse,gerstgrasser2024modelcollapseinevitablebreaking,kazdan2025collapsethriveperilspromises}. Surrogates with stronger distribution assumptions including Gaussian distributions, Gaussian processes, and GMMs provide a theoretically interpretable and tightly controllable testbed whose generative nature closely mirrors that of LLMs \citep{he2025goldenratioweightingprevents,suresh2024ratemodelcollapserecursive,borji2024noteshumailovetal,li2024preventingmodelcollapsegaussian,alemohammad2023selfconsuminggenerativemodelsmad}.

A crucial finding is that collapse dynamics shall be decisively influenced by the strategy for accumulating and infusing synthetic data across training iterations. A full replace paradigm \cite{schaeffer2025positionmodelcollapsedoes, dey2024universalitypi26pathwayavoiding} refers to entire updating in training data.
\citet{gerstgrasser2024modelcollapseinevitablebreaking} and \citet{kazdan2025collapsethriveperilspromises} validate that the accumulation of synthetic data from all previous iterations effectively prevents the collapse of the model.  \citet{bertrand2024stabilityiterativeretraininggenerative} only keep the most recent batch of synthetic data and points out synthetic to real data ratio shows a more significant role in model collapse. 

Recent work centers on establishing connection between model collapse and a varying synthetic-to-real ratio.  \citet{gillman2024selfcorrectingselfconsumingloopsgenerative} report that a moderate level of synthetic input helps preserve low-density diversity. \citet{he2025goldenratioweightingprevents} prove that, for both mean and covariance estimation, the estimation error is minimized when the weight on real data equals the reciprocal of the golden ratio. Conversely, \citet{dohmatob2024strongmodelcollapse} show that collapse can arise even at very small amount of synthetic data, implying that the synthetic-to-real ratio must asymptotically approach zero to maintain model diversity. Similarly, \citet{bertrand2024stabilityiterativeretraininggenerative} argue that keeping the ratio below a certain threshold also promotes long-run stability of training.

These results highlight the \emph{controllability} of model collapse.  
A systematic strategy for synthetic and real data combination shall lead to a framework of    interpretable model uncertainty control, more efficient use of synthetic data, and, cheaper but safer scaling of LLMs. Despite these promises, current research still faces challenges. \textbf{(i) Disconnect from realistic web dynamics and synthetic data generation}. Most existing analyses rely on purely statistical models, not tailored to mimic real-world Internet ecosystems. Additionally, their experimental setups often involve collecting and recycling synthetic generations across simulated time, a process that lacks a clear real-world analogue. \textbf{(ii) Computational limitations.} Full scale LLM model collapse experiments require iterative retraining and therefore remain extremely expensive, constraining empirical validation.

Our LWD framework is designed to bridge these gaps. We describe our experimental design in Section \ref{sec:methods}, and present empirical results in Section \ref{sec:results}, showing that the norm of the embedded distance matrix decreases during collapse in both LLM and GMM experiments. The limiting behavior of the GMM setup is proved in Appendix \ref{sec:app}. In Sections \ref{sec:discuss} and \ref{sec:limitations}, we discuss our findings and outline directions for future work.

%% file: text/methods.tex
\section{Methods and Experiments}\label{sec:methods}

\subsection{Experiment: Synthetic Conversations on the Internet}\label{sec:experiment_llm}

In our LLM Web Dynamics (LWD) framework, $n$ different LLM models  discuss a certain topic and get updated by answers from other participants over time. Full fine-tuning at each time is infeasible and expensive, so each model instead refreshes its context via retrieval-augmented generation (RAG).
RAG is lightweight to deploy and has been proven repeatedly showing equal or better performance in particular tasks comparing to  fine-tuning in empirical studies \citep{balaguer2024ragvsfinetuningpipelines,cheng2025surveyknowledgeorientedretrievalaugmentedgeneration,singh2025agenticretrievalaugmentedgenerationsurvey,gupta2024comprehensivesurveyretrievalaugmentedgeneration}, making it a practical and low-cost surrogate for iterative learning and training within the framework of LWD.

Suppose we have $n$ LLMs $\{\mathcal{M}_i\}_{i=1}^n$, each pre-trained on different data.  For instance, \texttt{GPT-3.5} is trained largely on English-dominant data \cite{brown2020languagemodelsfewshotlearners}, whereas \texttt{DeepSeek-67B} draws from a bilingual data of Chinese content \cite{deepseekai2024deepseekllmscalingopensource}.  These geographic and linguistic differences give the models heterogeneous priors over vocabulary, style, and factual knowledge, which is essential for studying cross-model information propagation \cite{alkhamissi2024investigatingculturalalignmentlarge,gupta2025multilingualperformancebiaseslarge}. To measure model performance, we use a fixed query $q$ and define a set of sentences containing information related to this query $q$. Intuitively, this set serves as the Internet which language models grab information from and contribute generated outputs to. In detail, at each time $t=1,2,\ldots$, each model augments itself by retrieving $k_t$ sentences from this set to generate a response to the query. Here we allow each model to grab a different number of sentences (and hence different amounts of information) as $t$ grows. In the standard RAG approach \citep{rag_survey}, LLM should retrieve the top $k_t$ relevant sentences; in our case, the sentences in the set are all relevant to $q$ by construction, so we draw $k_t$ sentences uniformly at random from this set each time.

We denote this set of sentences at time $t$ as $A^{(t)}$. If this set changes over time, each model's response would change accordingly, and thus we denote the response of model $i \in \{1,\ldots,n\}$ at time $t$ to query $q$ as $F_{i}^{(t)}(q)$.
For a fixed query $q$, there is still randomness in an LLM's output, so $F_{i}^{(t)}(q)$ is a random variable whose distribution depends on the model at time $t$ and the query. Our goal is to measure model performance, so at each $t$, we ask every model $L$ iterations to get $L$ responses $(f_{i}^{(t)}(q))_1,\cdots,(f_{i}^{(t)}(q))_L$ drawn independently from the distribution of $F_{i}^{(t)}(q)$.


At each time $t$, after all $n$ models each generate $L$ responses, we take one response from each model and ``post'' them to the ``Internet'' $A^{(t)}$. Under the i.i.d. assumption that we made previously, randomly picking one of $L$ responses is probabilistically equivalent to simply taking the first one. Thus, we can define the set $A$ at $t+1$ as:
$$A^{(t+1)}=A^{(t)}\cup\{(f_{1}^{(t)}(q))_{1},\cdots,(f_{n}^{(t)}(q))_{1}\}.$$
Hence $|A^{(t+1)}|=|A^{(t)}|+n$, reflecting that the Internet is increasingly filled with synthetic answers. If we begin with $|A^{(0)}|$ human-generated sentences, then at time $t$, the probability of retrieving a synthetic sentence from $A^{(t)}$ is $\frac{nt}{|A^{(0)}|+nt}$ (since we choose to retrieve them uniformly at random), which increasingly approaches $1$ as $t$ increases. Recall that at each $t$, each model fetches $k_t$ sentences from $A^{(t)}$. We want to use this experiment to discuss the case when LLMs get trained and updated by the information on the Internet, so it is reasonable to make $k_t$ proportional to $|A^{(t)}|$, i.e., $k_t=\lfloor \beta\cdot |A^{(t)}|\rfloor$ for a fixed hyperparameter $\beta$.


\begin{figure*}[t]
\centering
\includegraphics[width=\linewidth]{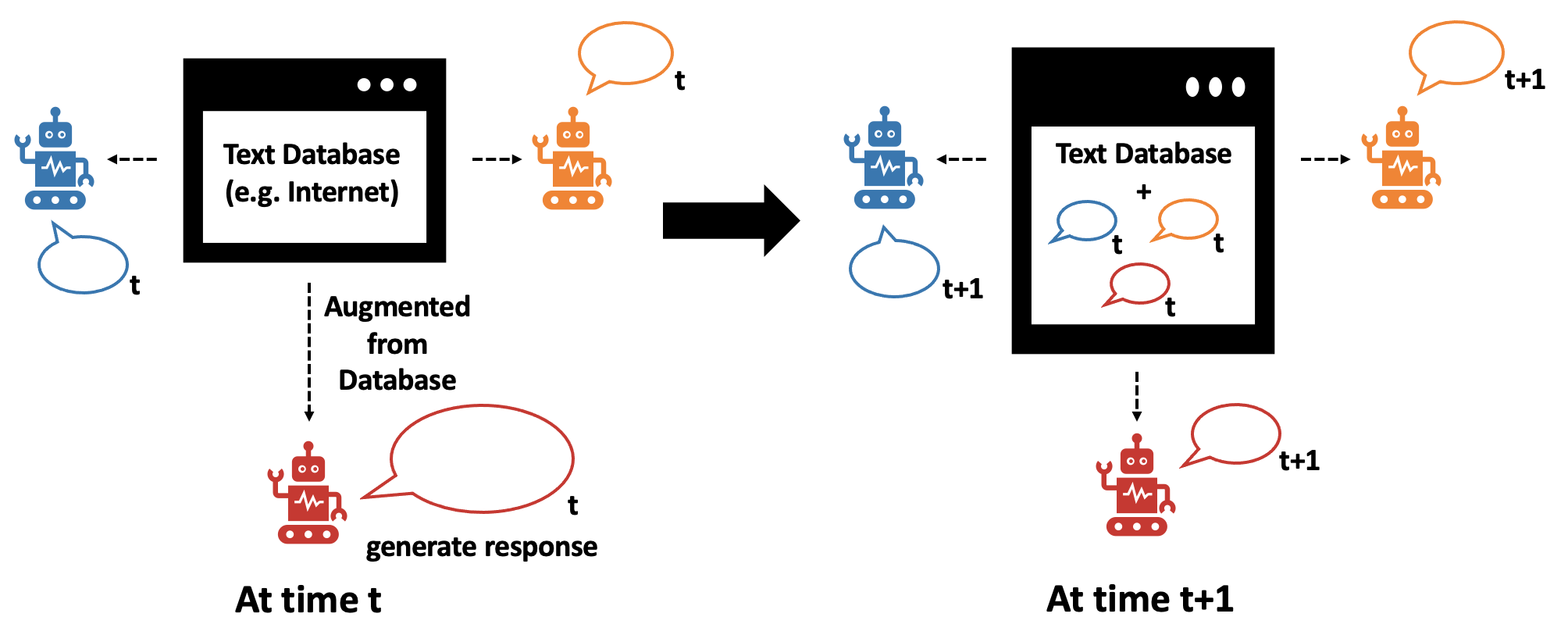}
\caption{Experiment in Section \ref{sec:experiment_llm}, from $t$ to $t+1$. At time $t$, three different LLMs generate responses based on augmentation contexts fetched from a shared text database. They then post their responses to the text database, based on which they generate the next-time responses.}
\label{fig:experiment}
\end{figure*}

Using this procedure (illustrated in Figure~\ref{fig:experiment}), we can observe how model responses evolve over time. To analyze the responses statistically, we use the open source embedding model \texttt{nomic-embed-v1.5} \citep{nomic}, denoted as $\phi$, which maps each text $(f_{i}^{(t)}(q))_l$ to a numeric vector $(\phi\circ f_{i}^{(t)}(q))_l\in\mathbb{R}^{768}$, $l=1,\cdots,L$. This function $\phi$ allows us to quantify the similarity between two texts such that a smaller distance between embedding vectors indicates that the original two sentences are more semantically similar. Given these $L$ vectors for each model $i$ at time $t$, we can estimate the distribution of $(\phi\circ F_{i}^{(t)}(q))$. This can be challenging due to the high dimension of the embedding model, and one can adopt the framework of \emph{Data-Kernel Perspective Space} (DKPS) \citep{helm2024tracking}, which reduces its dimension by Classical Multidimensional Scaling to get nice vector representations.

In addition to observing how the distribution evolves, we focus on estimating the distribution mean by averaging these $L$ vectors to measure how model $i$ responds to query $q$ at time $t$: 
\begin{equation}\label{eq:Xit}
X_{i}^{(t)} \coloneqq \frac{1}{L}\sum_{l=1}^L \left(\phi\circ f_{i}^{(t)}(q)\right)_l \in\mathbb{R}^{768}.
\end{equation}


We conjecture that as $t$ increases, the distributions of $F_i^{(t)}(q)$ and $F_j^{(t)}(q)$ for $i\neq j$ increasingly resemble each other, which would then imply the same behavior for the sample means $X_i^{(t)}$ and $X_j^{(t)}$. In detail, we capture the pairwise model difference through a $n\times n$ matrix $D^{(t)}$ whose entries are defined as
\begin{equation}\label{eq:Dt}
D^{(t)}_{ij}\coloneqq \left\|X_{i}^{(t)}-X_{j}^{(t)}\right\|_2,\ i,j=1,\cdots,n
\end{equation}
where $\|\cdot\|_2$ denotes the Euclidean norm. To discuss that model diversity shrinks under our experimental setting, we conjecture that $\left\|D^{(t)}\right\|_F$ converges to a small nonnegative number $c$ in probability, i.e.
$$\left\|D^{(t)}\right\|_F\xrightarrow{\ p\ } c \text{ as }t\to\infty,$$
where $\|\cdot\|_F$ denotes the Frobenius norm, and $c$ depends on some unknown intrinsic differences between models that are invariant to the environment. The detailed experiment is summarized in Algorithm~\ref{algo:llm}, and we present examples in Section~\ref{sec:results} to illustrate this conjecture.

\begin{algorithm*}
\caption{Experiment: synthetic conversations on the Internet}
\label{algo:llm}
\begin{algorithmic}[1]
\Require $n$ different LLMs, \texttt{nomic-embed-v1.5}, a set of texts $A^{(0)}$, a fixed query $q$, a fixed $L$, a fixed $\beta$, total time $T$.
\For{timestep $t=0$ to $T$}
    \For{each model $i=1$ to $n$}
        \State \textbf{Retrieve} $k_t$ sentences uniformly from $A^{(t)}$, where $k_t:=\lfloor \beta\cdot |A^{(t)}|\rfloor$.
        \State \textbf{Generate} answers based on the retrievals $L$ times: $(f_{i}^{(t)}(q))_1,\cdots,(f_{i}^{(t)}(q))_L$.
        \State \textbf{Embed} them: $(\phi\circ(f_{i}^{(t)}(q))_1,\cdots,(\phi\circ(f_{i}^{(t)}(q))_L$ using \texttt{nomic-embed-v1.5}.
        \State \textbf{Compute} $X_{i}^{(t)}$ using Eq.\eqref{eq:Xit}.
    \EndFor
    \State \textbf{Construct} matrix $D^{(t)}$ using Eq.\eqref{eq:Dt}. \textbf{Compute} $\|D^{(t)}\|_F$.
    \State \textbf{Update} the set $A$: $A^{(t+1)}=A^{(t)}\cup\{(f_{1}^{(t)}(q))_{1},\cdots,(f_{n}^{(t)}(q))_{1}\}$.
    \State \textbf{Get} $F_i^{(t+1)}(q)$ corresponding to the updated $A^{(t+1)}$, $i=1,\cdots,n$.
\EndFor
\end{algorithmic}
\end{algorithm*}

\subsection{A System of Gaussian Mixture Models}\label{sec:experiment_gmm}

Next we design a system of Gaussian mixture models (GMMs) analogous to the mechanism in Section~\ref{sec:experiment_llm}. Consider a collection of $n$ time-varying GMMs in $d$-dimensional Euclidean space. For all models, we fix the number of components $B$, a single nondegenerate covariance matrix $\Sigma\in \mathbb{R}^{d\times d}$, and mean vectors $\mu_b\in\mathbb{R}^d$, $b=1,\cdots,B$. 
We define the $i$-th GMM at $t$, denoted as $g_i^{(t)}$, via the distribution 
\begin{equation}\label{gmm_def}
g_i^{(t)}(x)=\sum_{b=1}^B \pi_{i,b}^{(t)}\mathcal{N}(x;\mu_b,\Sigma),
\end{equation}
where $\mathcal{N}(x;\mu_b,\Sigma)$ denotes the density of a normal distribution at $x$ with parameters $(\mu_b,\Sigma)$, and $\pi_{i,b}^{(t)}$ denotes the GMM's mixture weight at component $b$; 
we collect the $B$ mixture weights as the probability vector $\Pi_i^{(t)}$.
We initialize each GMM with different vectors $\Pi_i^{(t=0)}$. For example, one can sample each vector $\Pi_i^{(0)}$ from a Dirichlet distribution with parameter $(a, \ldots, a)$.

Recall that $\phi\circ (F_i^{(t)}(q))$ denotes the $i$-th LLM model at time $t$, evaluated by query $q$ and embedded by model $\phi$. It follows some distribution, which can be approximated by certain GMM like $g_i^{(t)}$. The parameters in $g_i^{(t)}$, including $\{B, \Sigma, \mu_b, \Pi_i^{(t)}\}$, can be all time-varying; for simplicity, we fix most of them and only allow $\Pi_{i}^{(t)}$ to update at each time. Hence, GMM $i$ at time $t$ is characterized by its mixture weights, the pairwise distance between $g_i^{(t)}$ and $g_j^{(t)}$ is reduced to the distance between $\Pi_{i}^{(t)}$ and $\Pi_{j}^{(t)}$, and we can define the $n\times n$ distance matrix at time $t$ as $D^{(t)}$:
\begin{equation}\label{eq:Dt_gmm}
(D^{(t)})_{ij}\coloneqq\left\|\Pi_{i}^{(t)}-\Pi_{j}^{(t)}\right\|_2.
\end{equation}

We then define a set of points $A$, similar to the set of texts (``the Internet'') in Section~\ref{sec:experiment_llm}. Let $A^{(0)}$ be the initialization of this set. At each $t$, each GMM $g_i^{(t)}$ generates $L$ points: $s^{(t)}_{i,1},\cdots,s^{(t)}_{i,L}$ to be added to the set $A$:
$$A^{(t+1)}=A^{(t)}\cup \bigcup_{i=1}^n\{s^{(t)}_{i,1},\cdots,s^{(t)}_{i,L}\}.$$

This set is used to update each GMM in the following way: each model $i$ randomly draws $k_t$ points from $A^{(t)}$. As in Section \ref{sec:experiment_llm}, we put $k_t=\lfloor \beta\cdot |A^{(t)}|\rfloor$. Using $\Pi_i^{(t)}$ as the prior probabilities and these $k_t$ points as observed data, we compute a posterior update of $\Pi_i^{(t+1)}$ via a revised update algorithm in \citet{zivkovic2004recursive}, as summarized in Algorithm~\ref{algo:gmm_update}:

\begin{algorithm}
\caption{Mixture-weight update algorithm}
\label{algo:gmm_update}
\begin{algorithmic}[1]
\Require Weight vector $\Pi_i^{(t)}=(\pi_{i,1}^{(t)},\cdots,\pi_{i,B}^{(t)})$, $k_t$ points $\{u_1,\cdots,u_{k_t}\}$, series of coefficients $\alpha_t\in(0,1)$ and $c_{t,B}$, threshold $\epsilon\in[0,1]$.
\State \textbf{Initialize} $\pi_{i,b}^{(t+1)} \leftarrow \pi_{i,b}^{(t)}$ for all $b=1,\ldots,B$.
\For{$u$ in $\{u_1,\cdots,u_{k_t}\}$}
    \State Compute the ownership $o_b=\pi_{i,b}^{(t+1)}\mathcal{N}(u;\mu_b,\Sigma)/g_i^{(t+1)}(u)$, $b=1,\cdots,B$; $g_i^{(t+1)}$ defined in Eq.\eqref{gmm_def}.
    \For{$b=1$ to $B$}
        \State $\pi_{i,b}^{(t+1)}\leftarrow(1-\alpha_t)\pi_{i,b}^{(t+1)}+\alpha_t\frac{o_b-c_{t,B}}{1-B\cdot c_{t,B}}$.
    \EndFor
    \For{$b=1$ to $B$}
    \If{$\pi_{i,b}^{(t+1)}\leq 0$}
        \For{$b' \neq b$}
            \State $\pi_{i,b'}^{(t+1)} \leftarrow \frac{\pi_{i,b'}^{(t+1)}}{1+\epsilon-\pi_{i,b}^{(t+1)}}$.
        \EndFor  
        \State $\pi_{i,b}^{(t+1)} \leftarrow \epsilon$.
    \EndIf
    \EndFor  
\EndFor
\end{algorithmic}
\end{algorithm}

The update of $\Pi_i^{(t)}$ thus updates $g_i^{(t)}$ to $g_i^{(t+1)}$. This experiment is summarized in Algorithm~\ref{algo:gmm}. For this system of GMMs, we conjecture that:
\begin{equation}\label{claim_gmm}
\|D^{(t)}\|_F\xrightarrow{\ p\ } 0 \text{ as }t\to\infty,
\end{equation}
Section~\ref{sec:results} presents simulation results supporting this conjecture, and we prove it in Appendix \ref{sec:app}.

\begin{algorithm*}
\caption{Simulation of a GMM system}
\label{algo:gmm}
\begin{algorithmic}[1]
\Require Number of models $n$, dimension $d$, number of components $B$, covariance matrix $\Sigma$, mean vectors $\{\mu_b\}_{b=1}^B$, positive integer $L$, fixed coefficient $\beta$, total time $T$.
\State \textbf{Initialize} mixture coefficients $\Pi_i^{(t)}\sim\text{Dir}(a)$ and $A^{(0)}$.
\For{timestep $t=0$ to $T$}
    \State \textbf{Construct} matrix $D^{(t)}$ using Eq.\eqref{eq:Dt_gmm}. \textbf{Compute} $\| D^{(t)} \|_F$.
    \For{each model $i = 1$ to $n$}
        \State \textbf{Generate} $L$ points $s^{(t)}_{i,1},\cdots,s^{(t)}_{i,L}$.
        \State \textbf{Sample} $k_t:=\lfloor \beta\cdot |A^{(t)}|\rfloor$ points uniformly from $A^{(t)}$, say, $\{ u^{(t)}_1, u^{(t)}_2, \dots, u^{(t)}_{k_t} \}$.
        \State \textbf{Update} $\Pi_i^{(t)}$ to $\Pi_i^{(t+1)}$ via Algorithm \ref{algo:gmm_update} using data $\{ u^{(t)}_1, u^{(t)}_2, \dots, u^{(t)}_{k_t} \}$. This also updates the GMM from $g_i^{(t)}$ to $g_i^{(t+1)}$.
    \EndFor
     \State \textbf{Update} $A$: $A^{(t+1)}=A^{(t)}\cup \bigcup_{i=1}^n\{s^{(t)}_{i,1},\cdots,s^{(t)}_{i,L}\}$.
\EndFor
\end{algorithmic}
\end{algorithm*}

%% file: text/results.tex
\section{Results}\label{sec:results}

\subsection{LLM Experiment}

To implement the experiment designed in Section \ref{sec:experiment_llm}, we try $n=3$ models: Meta's \texttt{Llama-3.1-8B-Instruct} \citep{dubey2024llama}, DeepSeek's \texttt{deepseek-llm-7b-chat} \citep{deepseek}, and Mistral's \texttt{Mistral-7B-Instruct-v0.3} \citep{mistral7b}. These three AI companies are headquartered in the U.S., China, and France respectively; we choose them so that the three models are pre-trained differently at least due to the distinct emphasis on contexts in their own languages. At $t=0$, we construct the ``Internet'' $A^{(0)}$ by putting 20 posts about cryptocurrencies, collected and organized by \citet{data_pejman}: 7 are positive, 7 are neutral, and 6 are negative. The query we use is \textit{``Please provide EXACTLY one concise sentence about the future prospects of Bitcoin.''}, and each model answers the question after reading posts sampled from $A^{(t)}$. We instruct them to answer exactly in one sentence because we want to focus on the semantic similarity or difference between these three models over time, instead of the language structure and complexity difference. The latter is also interesting to explore, but requires other statistical tools on texts, such as the depth of parse tree of sentences. We aim to investigate this in our future work.

\begin{figure}[t]
  \includegraphics[width=\columnwidth]{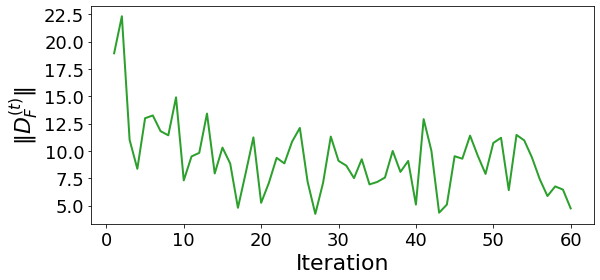}
  \caption{As $t$ approaches 60, the norm of the distance matrix in our LLM experiment, i.e. $\left\|D^{(t)}\right\|_F$ as defined in Eq.\eqref{eq:Dt}, decreases from approximately 20 to around 5.}
  \label{fig:llm_norm}
\end{figure}

We then follow Algorithm \ref{algo:llm} to simulate the experiment for $T=60$. We set $L=40$ and $\beta=0.5$. The experiment was conducted on \texttt{Nvidia A100 80GB GPUs, AMD EPYC 7443 24 Core}, taking approximately 8 hours to complete the full process. Figure \ref{fig:llm_norm} shows that the norm decreases from over 20 to around 5, indicating that the average responses generated by these three models are becoming more similar semantically.

In fact, this phenomenon is evident when directly comparing posts generated by the three models at the initial time and at $T=60$ (see Appendix \ref{sec:app2}). We can see that the models initially express opinions from different perspectives, but by $T=60$, their outputs become nearly identical semantically.

\begin{figure}[t]
  \includegraphics[width=\columnwidth]{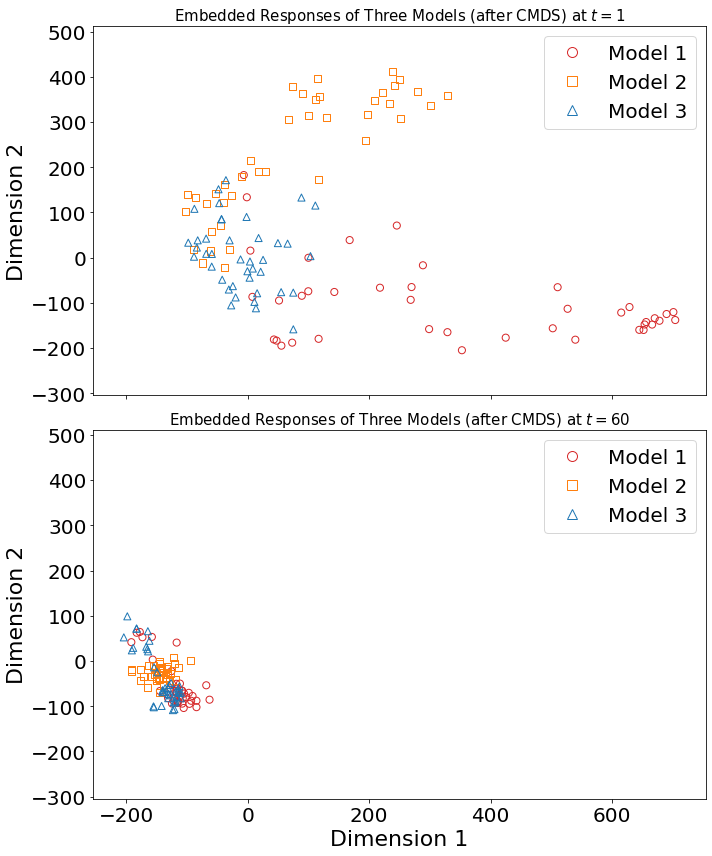}
  \caption{Scatter plots of the embedded responses from three LLMs in a 2D space at $t=1$ and $t=T$, with small random jitter added to reduce overlap and improve visibility. At $t=1$, the models exhibit clearly distinct response distributions. By $t=T$, the responses become tightly clustered and more similar across models.}
  \label{fig:llm_scatter}
\end{figure}

Recall that at each time $t$, we generate $L=40$ responses for each model so that we can examine their distributions. Since each embedded response is a vector of length 768, we perform Classical Multidimensional Scaling on the pairwise distance matrix so that we can plot each response in a 2-D space. From Figure \ref{fig:llm_scatter}, we can see that at the beginning, the distributions of three models' responses look very different, but at $T$, they are all point masses that are close to each other. This indicates that the three models are not only less diverse from each other, but also exhibit reduced internal diversity within themselves in the simulated Internet environment populated with synthetic posts.

\subsection{GMM Experiment}

Next we present some results of our GMM experiment described in Section \ref{sec:experiment_gmm} to demonstrate how it's phenomenally similar to the LLM experiment. First, we define three two-component GMMs ($B=2$). The component means are $\mu_1=-5,\ \mu_2=5$, and the variances are both $\Sigma_1=\Sigma_2=1$. The initial weights are sampled from Dirichlet with $a=1$, and same as the LLM experiment, we set $\beta=0.5$. As suggested by \citet{zivkovic2004recursive}, we set $\alpha_t = k_t^{-1}$; since $B\cdot c_{t,B}$ should be smaller than 1, we take $c_{t,B}=1/(k_t\cdot B^2)$. We take $L=3$ and $T=200$.

\begin{figure}[t]
  \includegraphics[width=\columnwidth]{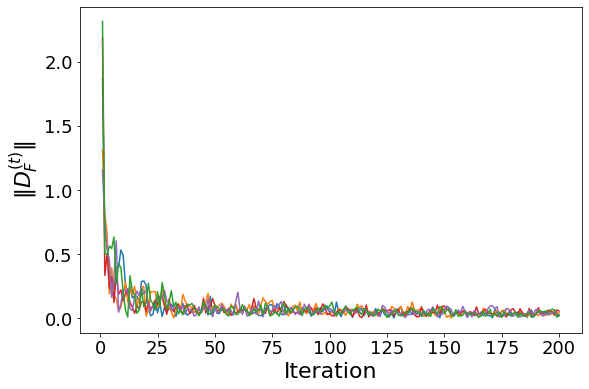}
  \caption{As $t$ increases, the norm of the distance matrix in our GMM experiment (with $B=2$), i.e. $\left\|D^{(t)}\right\|_F$ in Eq.\eqref{eq:Dt_gmm}, decreases to near zero across all 5 replicates.}
  \label{fig:gmm_2c_norm}
\end{figure}

Figure \ref{fig:gmm_2c_norm} shows how $\left\|D^{(t)}\right\|_F$ evolves in 5 replicates (i.e. by running the simulation 5 times with different seeds) - they all seem to converge to 0.

\begin{figure}[t]
  \includegraphics[width=\columnwidth]{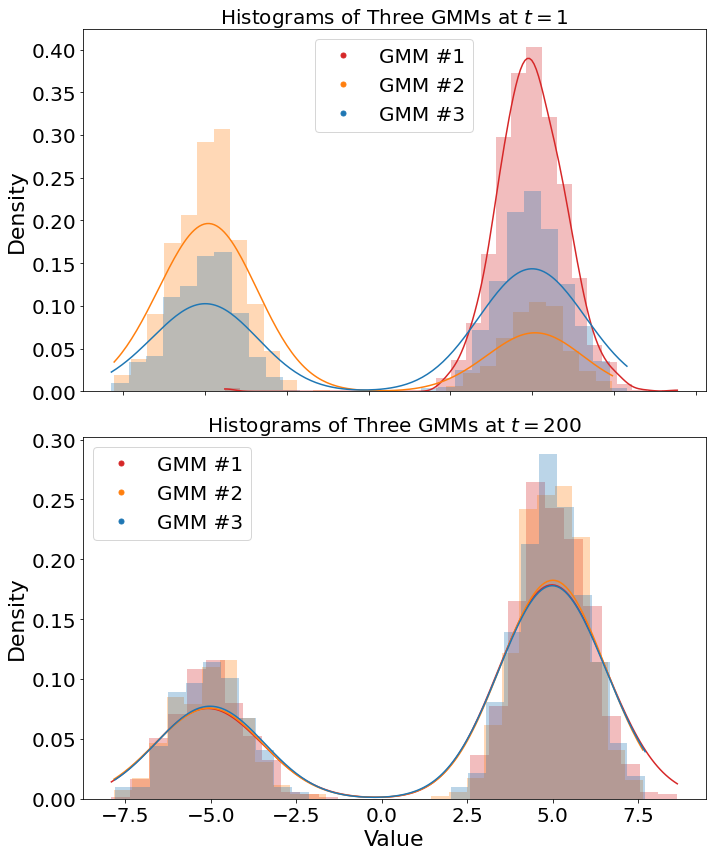}
  \caption{Histograms with density curves for three GMMs. At $t=1$, the distributions are clearly distinct, while at $t=T$, the density curves nearly coincide.}
  \label{fig:gmm_2c_histogram}
\end{figure}

Since GMM has an analytical form of density, we plot histograms of our three GMMs at different times. From Figure \ref{fig:gmm_2c_histogram}, we can see that they start with very different component weights but seem to converge to a same weight vector.

Indeed, we prove that for $i,j\in\{1,\cdots,n\}$, $\Pi_i^{(t)}-\Pi_j^{(t)}$ converges to 0 in expectation. From this we can further prove our claim (\ref{claim_gmm}) that $\|D^{(t)}\|_F$ converges to 0 in probability. We present the sketch of the proof in Appendix \ref{sec:app}.

When comparing the simulation results of our LLM and GMM experiments, we observe similar phenomena. Prior work, such as \citet{shumailov2024}, has also drawn parallels between LLM experiments and GMM setups. As a result, establishing theoretical claims in the GMM setting offers theoretical support for the behaviors observed in our LLM experiments.

However, we observe that the pattern of diminishing norms in Figure \ref{fig:gmm_2c_norm} looks smoother than that of LLMs in Figure \ref{fig:llm_norm}. This is because two-component GMM is much simpler than LLM. Using GMMs with a larger number of components serves as a better proxy for LLMs, as the additional components offer greater capacity to capture complex distributions.

\begin{figure}[t]
  \includegraphics[width=\columnwidth]{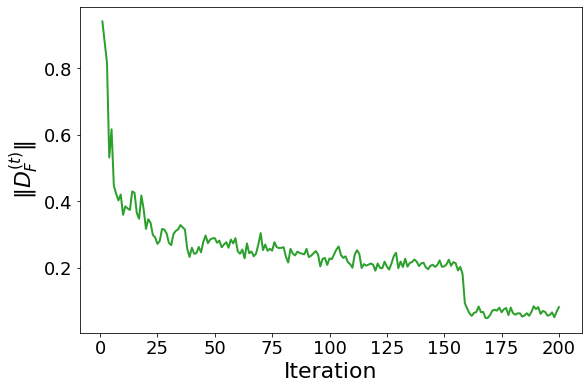}
  \caption{As $t$ increases, the norm of the distance matrix in our GMM experiment (with $B=11$) exhibits a decreasing trend, although the trajectory appears less smooth compared to Figure \ref{fig:gmm_2c_norm}. Notably, the norm continues to decline at some large $t$.}
  \label{fig:gmm_11c_norm}
\end{figure}

Figure \ref{fig:gmm_11c_norm} gives an example of our experiment using three GMMs of $B=11$ components. $\left\|D^{(t)}\right\|_F$ also seems to converge to 0, but the line looks less smooth and closer to the LLM plot. Interestingly, we see a sudden fall of $\left\|D^{(t)}\right\|_F$ around $t=160$ in Figure \ref{fig:gmm_11c_norm}. Therefore, it is highly likely that in our LLM experiment, the norm of the distance matrix between models will continue to decrease at some larger $t$. This highlights another advantage of our GMM setup—it serves as a useful proxy by offering insights into potential behaviors in the LLM experiment.

%% file: text/discussion.tex
\section{Discussion}\label{sec:discuss}

Our experiments demonstrate that model collapse can be observed and quantified in a fully API-accessible LLM network with RAG engine that mimics real-world Internet usage. By embedding individual responses and tracking the variation in induced Frobenius norm of the pairwise distance matrix, we observe a norm decay toward a small value, indicating that all agents have converged to an information neutral equilibrium. Importantly, this convergence should not be viewed as intrinsically \emph{good} or \emph{bad}; it merely signifies that further training using synthetic data shall no longer alter communication dynamics. Essentially the network used up all the information entropy from the raw data and communication built on the network is stabilized. Whether such “neutral” collapse is detrimental depends on goals of downstream tasks.

Our GMM setup within the LWD framework exhibits limiting behavior that closely mirrors that of our LLM experiment. One minor difference is the limiting value: we prove that the distance matrix norm in the GMM setting converges to zero in expectation, whereas in the LLM network, we only conjecture convergence to a small, non-negative value. The discrepancy arises because a $B$-component GMM is restricted to a finite parametrization, while a Transformer-based LLM occupies an effectively infinite-dimensional function space \cite{pérez2019turingcompletenessmodernneural,yang2020tensorprogramsiineural}. 

This distinction underscores a key advantage of our GMM proxy: its extensibility via changing parameterization. As demonstrated in Section \ref{sec:results}, GMMs with more components produce convergence patterns that more similar to those observed in real LLMs, illustrating the framework’s improved fidelity over simpler statistical surrogates. Moreover, in our setup, the means and variances of the Gaussians are fixed, and only the mixture weights are updated. These parameters can also be made time-varying, providing additional flexibility in approximating LLM dynamics. Importantly, simulating interactions in such complicated GMM system requires only minutes of computation, offering a highly efficient alternative for studying synthetic data pipelines before committing to expensive LLM experiments.

%% file: text/limitations.tex
\section{Limitations}\label{sec:limitations}
Despite the promise of our LLM Web Dynamics, several limitations remain. Our current study focuses on \emph{pattern learning} where we characterize collapse by geometric drift in terms of the diminishing norm of pairwise distance matrix in the embedded vector space.  
Because LLMs and their interaction graphs are essentially black-box models, these results could also be better justified by \emph{further statistical inference}. Therefore a key next step is to transform the pattern learning into systematic hypothesis testing or inference framework, providing finite-sample, high-probability guarantees. Finally, the LWD with RAG engine serves as a controllable testbed to quantify, isolate and exam the effect of different variables during the process of model collapse. Covariate like synthetic-to-real ratio, the distribution and quality of real data, and possible irrelevant noisy data perturbation can be incorporated into the framework LWD and produce predictable model collapse for various downstream tasks.

%% file: text/appendix.tex
\section{Sketch of Proof for Conjecture (\ref{claim_gmm})}\label{sec:app}

In this section, we discuss our conjecture (\ref{claim_gmm}) mathematically. In the GMM experiment, we measure the GMMs' weight differences in the distance matrix, so we start with proving pairwise weight differences going to 0.

For now, suppose we do not do the positive normalization step in Algorithm \ref{algo:gmm_update}. Let us consider the update of GMM $i=1,\ldots,N$ with points $u_1^{(t)}, \ldots, u_{k_t}^{(t)}$ sampled uniformly at random without replacement from $A^{(t)}$ (we denote this distribution as $\mathcal{U}(A^{(t)})$). Let us also consider the update for just one $u_j^{(t)}$ (i.e., one iteration of the outer for-loop in Algorithm~\ref{algo:gmm_update}). 
The mixture-weight update for component $b=1,\ldots,B$ is then
\begin{align}
\pi_{i,b}^{(t+1)}\leftarrow(1-\alpha_t)\pi_{i,b}^{(t)}+\frac{\alpha_t}{1-Bc_{t,B}}\left\{ o_{i,b}^{(t)}(u_j^{(t)})-c_{B}\right\}. 
\label{eq:weightupdate2}
\end{align}

Let us integrate the RHS over $u_j^{(t)} \sim \mathcal{U}(A^{(t)})$ for some arbitrary $j \in \{1,\ldots,k_t\}$.
For mathematical simplicity, suppose $A^{(0)}$ is empty. 
Under this simplifying assumption, the set $A^{(t)}$ contains only points generated from a (possibly previous) GMM, so the update point $u_j^{(t)}$ must have been generated by some GMM in an earlier step. Because the $k_t$ update points $u_1^{(t)}, u_2^{(t)}, \dots, u_{k_t}^{(t)}$ are drawn simultaneously from the same distribution, the order in which an update point is drawn does not affect the probability of which GMM (among $\{(i', t')\colon i' \in [n], t' \in [t-1]\}$) the point is drawn from.
Hence, we can say that $u_j^{(t)}$ is drawn from GMM $g_{i'}^{(t')}$ with probability $p_{i'}^{(t')} = (nt)^{-1}$.
We then integrate the ownership $o_{i,b}^{(t)}(u_j^{(t)})$, which is the only
term in the RHS of \eqref{eq:weightupdate2} that involves the update point:
\begin{align}\label{eq:ownership2}
    \int &o_{i,b}^{(t)}(u) \,\mathrm{d} g_{i'}^{(t')}(u)
\equiv \int\frac{\pi_{i,b}^{(t)} N(u;\mu_b, \Sigma)}{g_{i}^{(t)}(u)} \mathrm{d} g_{i'}^{(t')}(u)
\\&= \pi_{i,b}^{(t)} \int\frac{g_{i'}^{(t')}(u)}{g_{i}^{(t)}(u)}  \mathrm{d} N(u;\mu_b, \Sigma)
\end{align}
where $g_{i}^{(t)}(u)=\sum_{\ell=1}^{B}\pi_{i,\ell}^{(t)} N(u; \mu_b, \Sigma)$ is the likelihood of GMM $i$ at $u$. 
Because the GMM component parameters are fixed (only the mixture weights vary), if the mixture variances are small enough (almost same as saying if the mixture components are well-separated enough), then we can make the approximation $\frac{g_{i'}^{(t')}(u)}{g_{i}^{(t)}(u)} \approx \sum_{\ell=1}^B \frac{\pi_{i',\ell}^{(t')}}{\pi_{i,\ell}^{(t)}} 1_{(u = \mu_{\ell})}$. We also have that $N(u;\mu_b, \Sigma)$ is almost zero if $u$ is far enough away from $\mu_b$, so we can approximate $$\int\frac{g_{i'}^{(t')}(u)}{g_{i}^{(t)}(u)}  N(u;\mu_b, \Sigma) \mathrm{d} u  \approx \frac{\pi_{i',b}^{(t')}}{\pi_{i,b}^{(t)}}$$ 
so then Eq.\eqref{eq:ownership2} is approximated as $\pi_{i',b}^{(t')}$. Under this assumption, the integral of the RHS of Eq.\eqref{eq:weightupdate2} over $u_j^{(t)} \sim \mathcal{U}(A^{(t)})$
is approximately 
\begin{align} \label{eq:integralofRHS}
    \approx (1-\alpha_t)\pi_{i,b}^{(t)} + \alpha_t \frac{\bar\pi_b^{(t)} - c_{t,B} }{1-Bc_{t,B}} 
\end{align}
where we introduce the notation $$\bar\pi_b^{(t)} \coloneqq \sum_{i'=1}^n \sum_{t'=0}^{t-1} p_{i'}^{(t')} \pi_{i',b}^{(t')} = (nt)^{-1} \sum_{i'=1}^n \sum_{t'=0}^{t-1} \pi_{i',b}^{(t')},$$ which is component $b$'s mixture weight averaged over all $n$ models and all past times $0,\ldots,t-1$. 
If $0 < c_{t,B} < \min\{B^{-1}, \bar\pi_b^{(t)}\}$, then $\frac{\bar\pi_b^{(t)} - c_{t,B}}{1-Bc_{t,B}}$ in \eqref{eq:integralofRHS} is positive, which guarantees that the entire term in \eqref{eq:integralofRHS} is also positive.

If we do this update for each of the $k_t$ points, the ultimate update (in expectation) for $\pi_{i,b}^{(t+1)}$ is
\begin{align*}
    &\approx \bar\alpha_t^{k_t} \pi_{i,b}^{(t)} 
    + \alpha_t \frac{\bar\pi_b^{(t)} - c_{t,B}}{1-Bc_{t,B}} \sum_{s=0}^{k_t-1} \bar\alpha_t^{s} \\
    &= \bar\alpha_t^{k_t} \pi_{i,b}^{(t)}
    + \{1 - \bar\alpha_t^{k_t+1}\} \bigg[\frac{\bar\pi_b^{(t)} - c_{t,B}}{1-Bc_{t,B}}\bigg]
\end{align*}
where we introduce the notation $\bar\alpha_t = 1 - \alpha_t$.
Thus, at each $t$, the updated mixture-weight $\pi_{i,b}^{(t+1)}$ will, in expectation, be a convex combination of the (scaled) mixture weight $\pi_{i,b}^{(t)}/(1 - \alpha_t)$ and the value in the square brackets, which does not depend on which of the $n$ GMMs is being updated.
In particular, for any $i$ and $i'$ where $i \neq i'$, we have 
\begin{align*}
    E_t\left[\pi_{i,b}^{(t+1)} - \pi_{i',b}^{(t+1)}\right]
    &\approx \left(1-\alpha_t\right)^{k_t} \left[\pi_{i,b}^{(t)} - \pi_{i',b}^{(t)}\right].
\end{align*}
where $E_t$ denotes the conditional expectation w.r.t.\ the distribution described in Algorithms 2 and 3 given all quantities from all times up to and including time $t$.
By iterating over conditional expectations, we get
\begin{align*}
    E_0\left[\pi_{i,b}^{(t+1)} - \pi_{i',b}^{(t+1)}\right]
    &\approx \left[\prod_{s=0}^t \left(1-\alpha_s\right)^{k_s}\right] \left[\pi_{i,b}^{(0)} - \pi_{i',b}^{(0)}\right].
\end{align*}

If $\alpha_t = k_t^{-1}$, which is suggested by \citet{zivkovic2004recursive}, then because $\lim_{k_t \to \infty} \left(1-\alpha_t\right)^{k_t} = e^{-1}$ is a nonnegative constant strictly smaller than 1 and $\lim_{t \to \infty} k_t = \infty$, we conclude that $\lim_{t \to \infty} E_0 [\pi_{i,b}^{(t)} - \pi_{i',b}^{(t)}] = 0$ and thus also
\begin{align} \label{eq:distance-single}
    \lim_{t \to \infty} E_0 |\pi_{i,b}^{(t)} - \pi_{i',b}^{(t)}| = 0.
\end{align}

Note that
\begin{align*}
    E_0\left[\|D^{(t)}\|_F^2\right] 
    &= E_0\left[\sum_{i,i'} \left|D_{i,i'}^{(t)}\right|^2\right] \\
    &= E_0\left[\sum_{i,i'}^n \sum_{b=1}^B \left|\pi_{i,b}^{(t)} - \pi_{i',b}^{(t)}\right|^2\right] \\
    &= \sum_{i,i'}^n \sum_{b=1}^B E_0\left[\left|\pi_{i,b}^{(t)} - \pi_{i',b}^{(t)}\right|^2\right] \\
    &\leq \sum_{i,i'}^n \sum_{b=1}^B E_0\left|\pi_{i,b}^{(t)} - \pi_{i',b}^{(t)}\right|
\end{align*}
where the inequality comes from the fact that each mixture weight is a value between 0 and 1.
Thus, \eqref{eq:distance-single} implies \[\lim_{t \to \infty} E_0\left[\|D^{(t)}\|_F^2\right] = 0.\]
Because $E[X^2] \geq E^2[X]$ for any random variable $X$ and expectation $E$, we also get \[\lim_{t \to \infty} E_0^2\left[\|D^{(t)}\|_F\right] = 0\]
and thus also  \[\lim_{t \to \infty} E_0\left[\|D^{(t)}\|_F\right] = 0.\]
By Markov's inequality, the preceding panel implies that $\|D^{(t)}\|_F$ converges to 0 in probability.

Here we see that we get convergence even when we are still conditioned on the initial mixture-weight values. That is, we get convergence regardless of the starting mixture-weight values, other than they should be strictly between 0 and 1. Hence, we do not need any distributional assumptions on these starting weights $\Pi_i^{(t=0)}$.

%% file: text/appendix2.tex
\section{Example Posts Generated by Three Models at t=1 and t=60}\label{sec:app2}

\begin{itemize}
\item At $t=1$, \texttt{Llama-3.1-8B-Instruct}: \textit{Bitcoin's price is expected to remain stable around \$30,000 as the market awaits the Federal Reserve's announcement on interest rates despite recent fluctuations in global hash rate and market volatility.}
\item At $t=1$, \texttt{deepseek-llm-7b-chat}: \textit{Bitcoin's future prospects are expected to continue gaining mainstream adoption, with major institutions like PayPal, JPMorgan, and Tesla integrating crypto into their services and products.}
\item At $t=1$, \texttt{Mistral-7B-Instruct-v0.3}: \textit{The future prospects of Bitcoin continue to evolve, with increasing institutional adoption, regulatory decisions, and the emergence of new stablecoins and CBDCs impacting its growth potential.}
\item At $t=60$, \texttt{Llama-3.1-8B-Instruct}: \textit{The future prospects of Bitcoin are marked by growing mainstream adoption, institutional integration, increased user base, and regulatory decisions, while facing market volatility and security concerns.}
\item At $t=60$, \texttt{deepseek-llm-7b-chat}: \textit{The future prospects of Bitcoin indicate continued mainstream adoption, influenced by increasing institutional integration, a growing user base, and regulatory decisions, while facing market volatility and security concerns.}
\item At $t=60$, \texttt{Mistral-7B-Instruct-v0.3}: \textit{The future prospects of Bitcoin suggest continued mainstream adoption, driven by increasing institutional integration, a growing user base, and regulatory decisions, while facing market volatility and security concerns.}
\end{itemize}